\documentclass[10pt,twocolumn,letterpaper]{article}

\usepackage{cvpr}              % To produce the CAMERA-READY version
\usepackage{graphicx}
\usepackage{amsmath}
\usepackage{amssymb}
\usepackage{booktabs}
\usepackage{bbding}
\usepackage{multirow}
\usepackage{url}
\usepackage[symbol]{footmisc}
\usepackage[pagebackref,breaklinks,colorlinks]{hyperref}
\usepackage[capitalize]{cleveref}
\crefname{section}{Sec.}{Secs.}
\Crefname{section}{Section}{Sections}
\Crefname{table}{Table}{Tables}
\crefname{table}{Tab.}{Tabs.}

\def\cvprPaperID{4732} % *** Enter the CVPR Paper ID here
\def\confName{CVPR}
\def\confYear{2022}

\begin{document}

\title{A Large-scale Comprehensive Dataset and Copy-overlap Aware Evaluation Protocol for Segment-level Video Copy Detection}

\author{\normalsize{Sifeng He\thanks{These authors contributed equally to this research.} ,  Xudong Yang$^*$,  Chen Jiang$^*$, Gang Liang, Minlong Lu, Wei Zhang, Tan Pan, Qing Wang, Furong Xu,} \\\normalsize{Chunguang Li, Jingxiong Liu, Hui Xu, Kaiming Huang, Yuan Cheng, Feng Qian\thanks{Corresponding author.} , Xiaobo Zhang$^\dag$, Lei Yang}\\
Ant Group\\
{\tt\small \{sifeng.hsf$^*$, jiegang.yxd$^*$, qichen.jc$^*$, youzhi.qf$^\dag$, ayou.zxb$^\dag$\}@antgroup.com}
% {\tt\small \{sifeng.hsf, jiegang.yxd, qichen.jc, lianggang.liang, ivy.zw, pantan.pt, wq176625, booyoungxu.xfr, \\\tt\small {lichunguang.lcg, jinxiong.ljx, cup.xuh, kaiming.huangkm, chengyuan.c, youzhi.qf, ayou.zxb, yl149505 }\}@antgroup.com}
}

\maketitle

%%%%%%%%% ABSTRACT
\begin{abstract}
   In this paper, we introduce VCSL (Video Copy Segment Localization), a new comprehensive segment-level annotated video copy dataset. Compared with existing copy detection datasets restricted by either video-level annotation or small-scale, VCSL not only has two orders of magnitude more segment-level labelled data, with 160k realistic video copy pairs containing more than 280k localized copied segment pairs, but also covers a variety of video categories and a wide range of video duration. All the copied segments inside each collected video pair are manually extracted and accompanied by precisely annotated starting and ending timestamps.
   Alongside the dataset, we also propose a novel evaluation protocol that better measures the prediction accuracy of copy overlapping segments between a video pair and shows improved adaptability in different scenarios. 
   By benchmarking several baseline and state-of-the-art segment-level video copy detection methods with the proposed dataset and evaluation metric, we provide a comprehensive analysis that uncovers the strengths and weaknesses of current approaches, hoping to open up promising directions for future works. 
   The VCSL dataset, metric and benchmark codes are all publicly available at \href{https://github.com/alipay/VCSL}{https://github.com/alipay/VCSL}.

\end{abstract}

%%%%%%%%% BODY TEXT
\section{Introduction}
\label{sec:intro}

In recent years, the wide spread of pirated multimedia has attracted attention from both users and platforms all over the world. The dramatic rise of pirated content has been fueled by the large amounts of user-generated content (UGC) and professionally-generated content (PGC) uploaded to content sharing market, e.g., over 500 hours of video are uploaded to YouTube every minute ~\cite{Youtube21} , the average monthly paying users of Bilibili have increased by 62\% in one year alone ~\cite{bilibili21}. These videos can generate significant advertising revenue, providing strong incentive for unscrupulous individuals that wish to capitalize on this bonanza by skillful copyright infringement ~\cite{xu2017caught}. Some video editing specialists even design methods to evade infringement detection algorithms by cropping, melting, and merging short clips from popular videos which make accurate copy detection even more challenging. Given the escalating adversarial relation between the platform algorithm and evolved piracy, comprehensive datasets with real partial video infringement become increasingly essential.

Besides copyright protection, a video copy detection (VCD) system is important in applications like video classification, tracking, filtering and recommendation ~\cite{law2007video,visil, chanussot2021}. In most cases, video-level copy detection results alone are not sufficient as the detected videos are usually displayed and interacted with system users for downstream tasks. Hence, designing an approach that can locate the copied segments is preferred and has already attracted lots of attentions in recent works ~\cite{lamv, spd, han2021video, tan2021fast, vcdb}.

However, manually annotating copied segments between videos is time-consuming and costly. Some datasets for copy detection, e.g., CCWEB ~\cite{ccweb}, FIVR ~\cite{fivr} and SVD ~\cite{svd} provide only video-level annotation indicating whether two videos contain copied parts or not, which is coarse-grained and incapable of evaluating segment-level copy detection methods. Other datasets, e.g., MUSCLE-VCD ~\cite{muscle_vcd} and TRECVID ~\cite{trecvid}, produce automatically segment-level labels by generating simulated copied segments with pre-defined transformations, which may not be representative for real-world data ~\cite{vcdb}. 
% Besides, the category and video length distribution of these datasets are lack of diversity. % Besides, these datasets are lack of category diversity and video length variations.
The only manually-labelled segment-level dataset, VCDB ~\cite{vcdb} released in 2014, contains only 6k labelled videos pairs with 9k segment pairs, and over 70\% of copy durations are less than 1minute. The annotation quantity and video diversity of existing datasets are not sufficient to develop segment-level video copy detection algorithms that need training data and labels. 

To address these issues, we present a comprehensive dataset, VCSL, specifically designed for segment-level video copy detection. This dataset, which will be made publicly available, contains over 160k infringed video pairs with 280k carefully annotated segment pairs. All of these videos are realistic copies from Youtube or Bilibili, which cover a wide range of video topics including movies, music videos, sports, etc.

Meanwhile, existing evaluation protocols for segment-level video copy detection exhibit an obvious drawback that most of them utilize ground-truth copied segments as queries rather than the entire videos ~\cite{vcdb,lamv,spd}. This is unpractical for real copy detection scenario where it is hard to know a priori that which part of a video will be pirated. Hence, we indicate a protocol that is more realistic with a pair of copied videos as input, and a new metric is jointly proposed to address previous unreasonable issues. With the awareness of both copied segments and overlap accuracy, our proposed metric fully considers the distinctiveness regarding the segment division equivalence (illustrated in Sec.4.1) of copy detection task, and it is more suitable and robust for various infringe situations.

Furthermore, we introduce a benchmark for segment-level video copy detection. We decouple the entire algorithm process into two main algorithm modules: feature extraction and temporal alignment. Then we evaluate the baseline and state-of-the-art (SOTA) algorithms of both parts on the splitted test set of VCSL. The components described above represent a complete benchmark suite, providing researchers with the necessary tools to facilitate the evaluation of their methods and advance the field of segment-level video copy detection.

\section{Related Work}
In this section we provide an overview of datasets and evaluation metrics designed for different video copy detection and retrieval tasks, followed by a survey of techniques targeting segment-level video copy detection.

\subsection{Datasets and Evaluation}
We briefly review the datasets for VCD task in this section including CCWEB ~\cite{ccweb}, MUSCLE VCD ~\cite{muscle_vcd}, TRECVID ~\cite{trecvid}, FIVR ~\cite{fivr}, SVD ~\cite{svd} and VCDB ~\cite{vcdb}.

CCWEB ~\cite{ccweb} dataset is one of the most widely used dataset. It contains 24 query videos and 12,790 labelled videos. 
% The videos were collected by submitting 24 popular text queries to video sharing websites. 
All retrieved videos in the video sets were manually annotated by three annotators based on their video-level relation to the query video. In addition to only video-level annotation, CCWEB also shows limitations on both video transformation and topic category diversity, and almost all the recent methods can achieve a near-perfect performance (video-level mAP of $>$0.99) on this dataset. 

MUSCLE-VCD ~\cite{muscle_vcd} collects 18 videos to construct query set. Then the authors utilize query videos to generate 101 videos as labeled set based on some predefined transformations. Similarly, TRECVID datasets ~\cite{trecvid} were constructed following the same process as the MUSCLE-VCD dataset. The latest edition of the dataset contains 11,503 reference videos of over 420 hours and 11,256 queries. The queries were automatically generated by randomly extracting a segment from a dataset video and imposing a few predefined transformations. Therefore, copies in MUSCLE-VCD and TRECVID are all simulated based on predictable processing, which are less diversified and much easier to be detected and retrieved.

FIVR ~\cite{fivr} consists of 225,960 videos and 100 queries. This dataset collects fine-grained incident retrieved videos including three retrieval tasks: a) the Duplicate Scene Video Retrieval (DSVR), b) the Complementary Scene Video Retrieval (CSVR), and c) the Incident Scene Video Retrieval (ISVR). Despite the large-scale video collection of FIVR, only part of the first task (about 1325 annotated videos) in this dataset is relevant to the scope of video infringement, and all the videos are news event labelled in video-level. Similarly, SVD ~\cite{svd} is also a large-scale near duplicated dataset with only video-level annotation, and most of videos in SVD are less than 20 seconds.

The most relevant dataset to our work is VCDB dataset ~\cite{vcdb} consisting of 28 query sets and 528 labelled videos with 9,236 pairs of copied segments. The annotation gives the precise temporal location of each copied pair and thus is suitable for segment-level copy detection task. However, a couple of obvious weaknesses exist in VCDB. First, the amount of both labelled video data and positive pairs are too limited for further usage. Some topic categories contain only one query set, and it is impossible to split train and test set from it. Over 90\% of the videos in VCDB is less than 3 minutes which is also lack of diversity on duration. 

Table 1 summarizes some statistics of the aforementioned datasets. Currently, there is no dataset that simultaneously supports segment-level annotation, real copies collection and large-scale diversity. This motivates us to build a comprehensive video partial copy dataset.

\begin{table*}[]
\caption{Comparison between VCSL and existing datasets. As we cannot access MUSCLE-VCD and TRECVID datasets, some statistics of these two datasets are N/A. The segment statistics (last two rows) of datasets with only video-level annotation (CCWEB, FIVR, SVD) are also N/A. ($^1$) denotes the total of near-duplicated, duplicated scene, complementary scene and incident scene labels in FIVR. ($^2$) indicates only near-duplicated video pair as copied videos. All the durations are calculated on labelled videos.}
    \centering
    \scalebox{0.85}{
    \begin{tabular}{c|c|c|c|c|c|c|c|c}
    \toprule
Item & CCWEB  & MUSCLE-VCD & TRECVID & FIVR & SVD & FIVR-PVCD & VCDB &\textbf{VCSL}\\ \toprule
Segment-level annotation & \XSolidBrush   & \CheckmarkBold   & \CheckmarkBold  & \XSolidBrush   & \XSolidBrush  & \CheckmarkBold  & \CheckmarkBold   & \CheckmarkBold \\ \hline
Type of copies   & Realistic   & Simulated   & Simulated   & Realistic   & Realistic  & Realistic  & Realistic  & Realistic  \\ \hline
\#query sets    & 24   & 18   & 11256   & 100   & 1206  & 100  & 28    & 122\\\hline
\#labelled videos    & 12,790   & 101   & 11,503   & 12,868$^1$   & 34,020 & 5,964 & 528    & 9,207 \\ \hline
Average duration (in second) & 151.02   & 3564.36   & 131.44   &  113.12  & 17.33 & 113.12   & 72.77    & 364.90  \\ \hline
\#positive video pairs & 3,481   & N/A   & N/A   & 1,325$^2$  &5935 & 10,211   & 6,139    & \textbf{167,508}\\ \hline
\#copied segments    & N/A   & N/A   & N/A   & N/A   & N/A    &10,870 & 9,236  & \textbf{281,182} \\ \hline
% \textbf{\#negative pairs} & 1.8750   & 15.00   & 74.34   & 1.8750   & 15.00   & 74.34    & 74.34 \\ \hline 
Total copied duration (in hours) & N/A   & N/A   & N/A   & N/A   & N/A   &  76.4 & 326.8   &  \textbf{17,416.2}  \\ \bottomrule
\end{tabular}}
    \label{tab:alignment}
\end{table*}

In the aspect of evaluation metric, the video-level evaluation metric (mAP) of VCD task has been well discussed in previous work~\cite{ccweb,visil} and will not be covered in this paper. Previous segment-level evaluation metrics are introduced with MUSCLE-VCD ~\cite{muscle_vcd} and VCDB datasets ~\cite{vcdb}. Most of recent research works ~\cite{lamv,spd, han2021video} adopt segment precision and recall defined in VCDB as follows:
\begin{equation}
\begin{aligned}
  SP &= \frac{\left | correctly\  detected\  segments \right |}{\left | all\   detected\  segments \right |}\\
SR &= \frac{\left | correctly\   detected\  segments \right |}{\left | groundtruth\  copy \  segments\right |}
  \label{eq1}
\end{aligned}
\end{equation}

In addition, VCDB also introduces a frame-level metric as auxiliary criteria: 

\begin{equation}
\begin{aligned}
  FP &= \frac{\left | correctly\   detected\  frames \right |}{\left | all\   detected\  frames \right |}\\
FR &= \frac{\left | correctly\   detected\  frames \right |}{\left | groundtruth\  copy \  frames\right |}
  \label{eq2}
\end{aligned}
\end{equation}

However, both segment P/R and frame P/R have their limitations. The most significant one is that the protocol uses each segment in copy pairs rather than entire video as a query. Meanwhile, for segment P/R metric, a detected segment pair is considered correct as long as both of them have at least one frame overlap with the ground truth pair, leading to a poor awareness of copy overlap and alignment accuracy. Hence, we need to unify the detection performance and alignment accuracy into one metric, and make it appropriate for different infringement scenarios.

\subsection{Methods}
Frame-level features are proved to gain a large margin in video retrieval tasks \cite{visil,tca} and are necessary to precisely locate copied segments.
%%For frame-level feature representation, 
Current methods employ Deep CNNs \cite{mac17} and Deep Metric Learning (DML) \cite{dml2017,tca} to extract robust features. The application of Maximum Activation of Convolutions (MAC) and its variants \cite{mac17,mac2} %%on feature maps of convolution layers of a CNN architecture
has proved to be an efficient representation in retrieval tasks. Recently, transformer ~\cite{transformer} has recently emerged as an alternative to CNNs for visual recognition task ~\cite{vit,dino}. Self-supervised pretrained transformer model shows competitive performance on image copy detection task ~\cite{dino}.

After obtaining frame-level feature representation, a temporal alignment module needs to reveal the similarity and time range of one or multiple copied segments between the potential copied video pair. A simple method is to vote temporally by Temporal Hough Voting \cite{voting1,vcdb}. The graph-based Temporal Network (TN) \cite{temporaNetwork} takes matched frames as nodes and similarities between frames as weights of links to construct a network, and the matched clip is the weighted longest path in the network. Another method is Dynamic programming \cite{dp1} to find a diagonal blocks with the largest similarity. Inspired by temporal matching kernel~\cite{Poullot2015}, LAMV\cite{lamv} transforms the kernel into a differentiable layer to find temporal alignments. SPD ~\cite{spd} formulates temporal alignment as an object detection task on the frame-to-frame similarity matrix, achieving a state-of-the-art segment-level copy detection performance.

\section{Dataset}

\subsection{Annotation}
The dataset is constructed for meeting the following requirements: 1) the video copy transformations should be as diverse as possible, but avoid excessive transformations leading to extremely low image quality. 
% Most copies maintain or even recreate the quality of original video in real scenario; 
2) the category should cover most of common video topics; 3) the duration of video should not be limited to only one type (short video or long video). Based on the above requirements, we start from 122 carefully selected seed videos containing both PGC and UGC from Youtube and Bilibili. Each seed video is associated with a text query (keywords), which will be used to search potential relevant videos online. These 122 queries contain 11 common topics, i.e., movies, TV series, music videos, sports, games, variety show, animation, daily life, advertisement, news and kichiku. According to these seed videos and their corresponding text queries, around 100 videos per query are collected as potential copied videos from Youtube and Bilibili platform. 

Different from previous video-level annotation, extracting and annotating copied segments from their parent videos is an extremely error-prone and sophisticated task, especially for some short copied clips in currently fashionable "kichiku" videos. Therefore, we design a coordination process between algorithm engineers (us) and annotators with multiple steps shown in Fig.1. In terms of annotation cost, we employed 30 full-time well-trained annotators and spent about 4 months to finish entire annotation process (about 20,000 man-hours). All the following annotation steps contain a round of labelling by one annotator, a round of quality check by another annotator and a final round of spot check by us.

\begin{figure}[ht!] % use float package if you want it here
  \centering
  \includegraphics[width=1\linewidth]{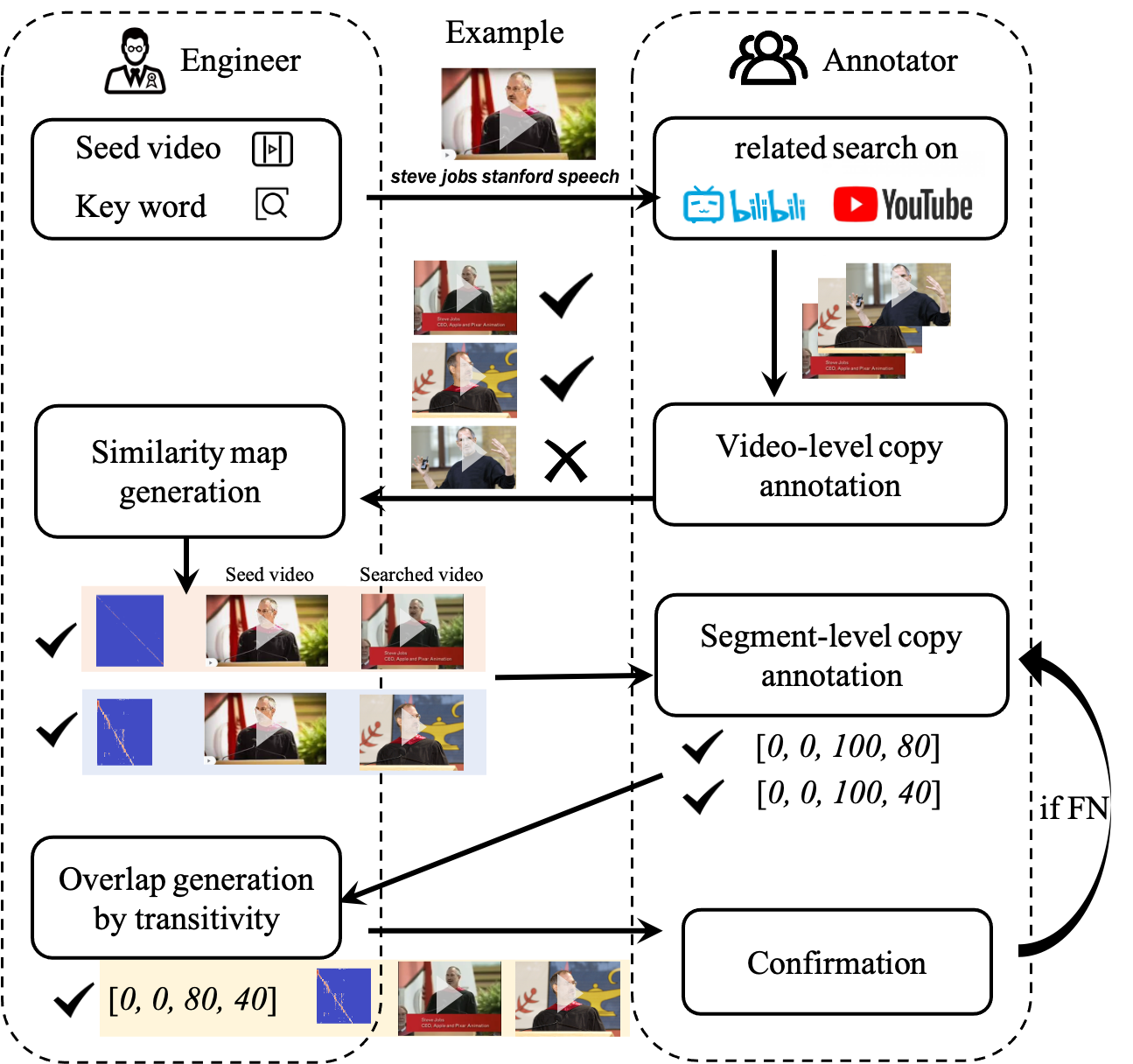}
  \caption{Overview of annotation process. Engineers (authors of this paper) in the left provide the initial query sets, build the annotation tool and clean the annotated data for next-step annotation. Annotators finish related search online, video-level and segment-level annotation. The similarity map is generated by dot production of frame-level features between the video pair. The segment annotation is expressed as a list with [\textit{start timestamp of seed video, start timestamp of searched video, end timestamp of seed video, end timestamp of searched video}].}
\end{figure}

As mentioned before, the first step of annotation is to provide the seed videos and text queries to all annotators, then require them to search the related videos and give us a coarse video-level copy result. The second step is segment-level annotation on the filtered copied videos in each query set from the first step. At this step, precisely locating the matched segment temporal boundaries is extremely time-consuming, and even the experienced annotator can only finish 2-3 video pairs within one hour. Here, we build an annotation tool, which shows not only the original video pairs but also the frame-to-frame similarity map (illustrated in Sec.S1 of Supplementary Material in detail) as auxiliary assist to annotators. By observing approximate straight lines in the similarity map ~\cite{spd}, the annotators can easily check missing copied clips after video comparison. After segment-level annotation, we obtain the segment copy information between one seed video and all the searched videos for each query set. Similar to VCDB datasets ~\cite{vcdb}, we utilize transitivity property of the video copies to automatically generate new copied segments between the searched videos if their matched segments in the seed video intersect. It is notable that annotating copied segments between two videos relevant to the same seed video with the transitivity property can bring both false positives (e.g. the copied segments share no common contents) and false negative (e.g. there exist copied segments not appeared in the seed video) annotations. Therefore, in the final step, the annotators first check if the copied segments from transitivity propagation are correct matches and refine the copy boundaries. Then they are provided with the two videos and the similarity map similar to that in the second step to find potential copied segments out of the remaining area. All the above annotation process is shown in Fig.1.

% image plot

\subsection{Statistics}

In total, we collect 9207 copied videos associated with 122 selected video queries from Bilibili and Youtube. After several rounds of carefully cooperated annotation by us and annotators, we extract and label 167,508 copied video pairs and 281,182 copied segments, both of which are two orders of magnitude more than the only realistic segment-level dataset VCDB. The detailed comparison between different datasets is shown in Table.1. It can be observed from the last row of Table.1 that the total copied duration of all segments in VCSL is even larger than total video duration in most of public available datasets, which shows the considerable large-scale of our dataset.

Fig.2 further shows some statistics of VCSL, and some detailed comparisons with VCDB. Different from VCDB and other short video datasets, VCSL contains videos longer than 30min, and these long videos include TV series and movies that are easy to be infringed nowadays. Meanwhile, the copied segment duration also covers a wider range from less than 5 seconds to even larger than 30min. Among the video pairs that have at least one segment copy, as high as 30\% of them contain two or more copied segments and 45\% of these segments are shorter than 1/5 of their parent video. All 122 video query sets are divided into 11 topics and the smallest topic contains at least three query sets for satisfying train-val-test splits. The least number of video copy pairs shown in Fig.2(e) for each topic is over 4k, which is more than half of labelled data (6k) from VCDB. Moreover, VCSL covers lots of realistic spatial and temporal transformations, and we list some in the Sec.S2 of Supplementary Material. The breadth and diversity of VCSL enable thorough comparisons between segment-level VCD approaches and make it possible to train supervised learning methods for which training data and labels are necessary.

\begin{figure}[ht!] % use float package if you want it here
  \centering
  \includegraphics[width=1\linewidth]{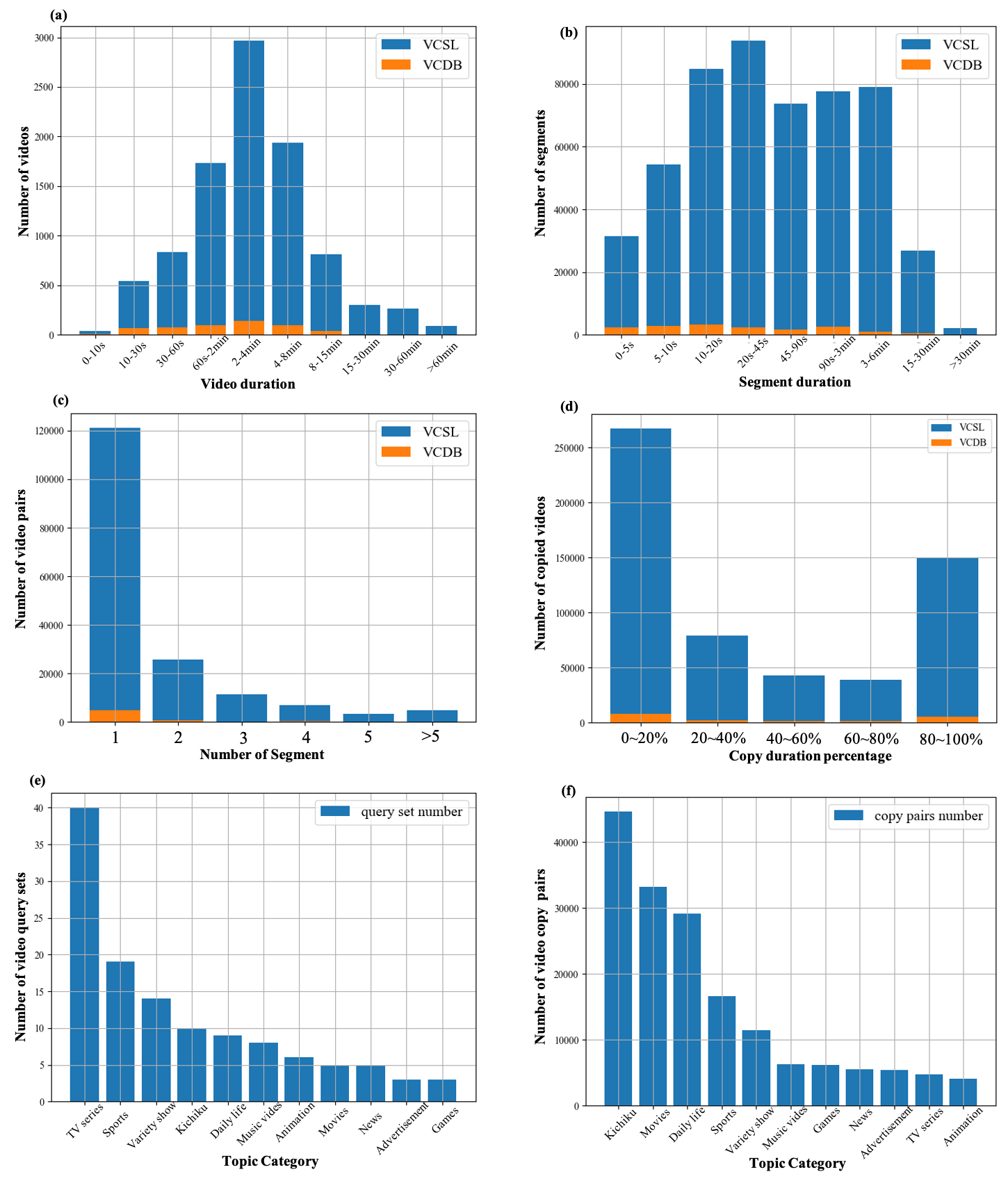}
  \caption{Data distribution of VCSL. All the blue bars represent the amount of VCSL and orange are for VCDB. (a) the numbers of videos in different video duration; (b) the numbers of segments in different segment duration; (c) the number of segments per video pair; (d) the duration percentage of the copy segments in their corresponding parent videos; (e) the number of video query sets for each topic category; (f) the number of video copy pairs for each topic category.}
\end{figure}

\section{Evaluation Protocol}
\subsection{Background and Motivation}
As mentioned before, previous protocols use only ground-truth copied segments as queries which is inappropriate for practical use. Instead, we refine the evaluation protocol by taking two entire videos as input and the system need to detect all the potential copied segments between the two videos. In this setting, most of previous metrics are inapplicable or need to be extended. As a result, we design a new metric to address this. 

However, the evolved evaluation protocol brings new difficulties for designing the metric. During the annotation process, we observe that the boundaries of copied segments are hard to determine in some cases. As an example shown in Fig.3(a), some intermediate frames are edited or briefly inserted by other video frames, leading to ambiguous segment boundaries. Other common cases are mashup videos shown in Fig.3(b). If one single entire copied segment pair and a sequence of consecutive sub-segment pairs occupy the same copied part on original video pairs, we believe that these two annotations are both reasonable and correct. This also applies to predictions of algorithms with different inductive biases. The equivalence of an entire copied segment pair and its division of consecutive sub-segment pairs, i.e., segment division equivalence, must be taken into account when designing the new metric.

We decide to use the precision and recall as the evaluation metric because they are widely adopted. But calculating the recall and precision similar to previous metrics in Eq.(1) and Eq.(2) is problematic. The segment-level precision and recall in Eq.(1) fail to measure the overlap of copied segments, and the frame-level precision and recall in Eq.(2) calculated respectively on two videos give totally wrong results in some cases which are demonstrated in Fig.4(f). A better way to evaluate the recall and precision in this scenario is explained in the next section.

\begin{figure*}
\begin{center}
\includegraphics[width=16cm,keepaspectratio]{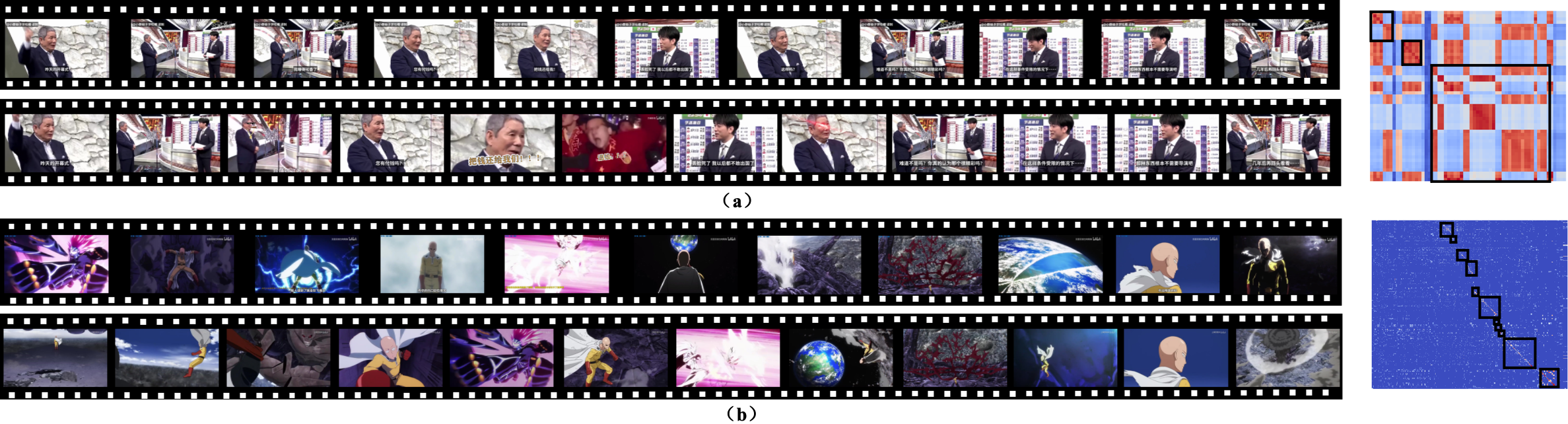}
\end{center}
   \caption{Two copied video examples with ambiguous segment boundaries. The left part is the timing screenshots of both videos, and right part is the similarity map between them with the most fine-grained copied segment annotations shown as black boxes. There are multiple boxes in frame-to-frame similarity map which can also be merged, and metric should not change dramatically whether these segment pairs are merged or not. These two similarity maps are quite different in appearance due to video duration and self-similarity between frames. }
\label{fig:short}
\end{figure*}

% On the other hand, only frame-level statistic Eq.(2) ignoring segment results is also not convinced. Firstly, previous frame-level precision and recall can only be calculated when the query input is a confirmed copied segment. Secondly, this frame granularity statistics cannot contain the temporal corresponding relations between the two compared video. In other words, this is only calculated on $x$ and $y$ axis in similarity map without correlation between them. Extreme case is the misalignment but accidentally similar projection on axis, which will be discussed later in Fig.4(f).

\subsection{New Metric}

%  The metric calculate the frame-level measurement inside each segment. 
 The calculation of the metric can be more clearly described on the frame-to-frame similarity map between a video pair shown in Fig.4. For ease of representation, all the copied segment correspondences are depicted as bounding boxes in Fig.4 (a-f), and the copied pattern in similarity map in Fig.4 (c-f) are shown in oblique straight lines which represent the temporal sequential copy between two videos. The predicted and ground truth segment pairs inside the video pair are respectively denoted as predicted bounding boxes $\left \{P_j\right \}_{1,2,...,n}$ and GT bounding boxes $\left \{G_i\right \}_{1,2,...,m}$ shown in Fig.4.

\begin{figure}[ht!] % use float package if you want it here
  \centering
  \includegraphics[width=1\linewidth]{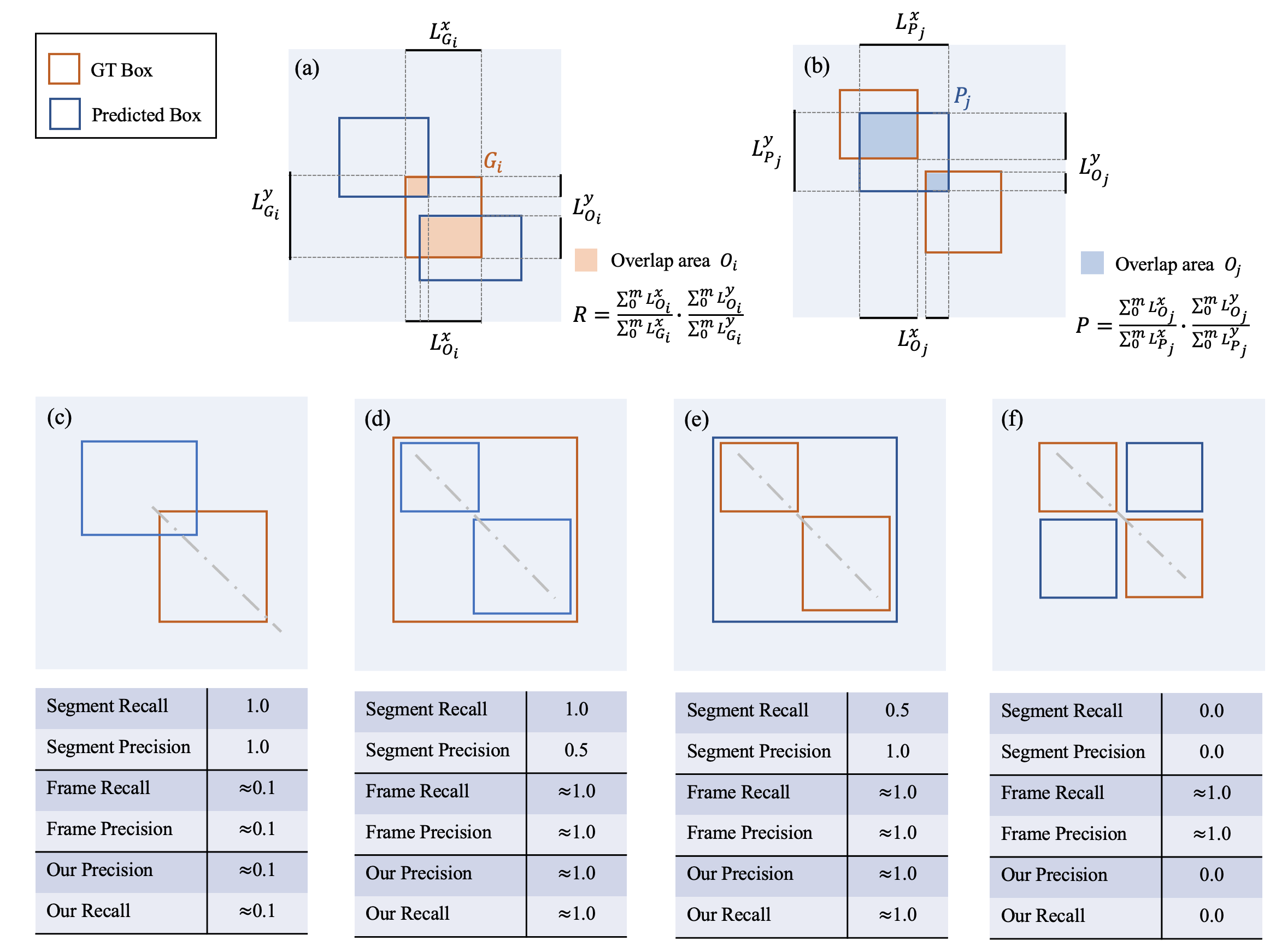}
  \caption{(a-b) illustrates calculation process of our proposed methods. (c-f) provide four simplified cases for comparison between our metric and extended segment and frame metrics. The gray dashed lines in (c-f) represent temporal sequential copy, and other common cases including random order video editing can also happen with sophisticated pattern in similarity map.}
\end{figure}

Specifically, we firstly define the overlapped regions between all the predicted bounding boxes $\left \{P_j\right \}_{1,2,...,n}$ and each GT bounding box $G_i$ with Intersect-over-Union (IoU) $>$ 0 as:
\begin{equation}
O_i = \left \{ P_1 \cap G_i, P_2 \cap  G_i, ... ,P_n \cap  G_i \right \}
  \label{eq3}
\end{equation}
Then for each GT box $G_i$, we calculate the union length $L_{O_i}^x$ and $L_{O_i}^y$ of projected lines (which indicate frames) from $O_i$ along both $x/y$ axis (which indicate temporal axis of video $A$ and video $B$). This process can be shown in Fig.4(a). We can also obtain the width and height of $G_i$ as $L_{G_i}^x$ and $L_{G_i}^y$.

Therefore, the recall metric of a video pair is defined as follows:
\begin{equation}
Recall = \frac{\sum_{i=0}^{m}L_{O_i}^x}{\sum_{i=0}^{m}L_{G_i}^x}\cdot \frac{\sum_{i=0}^{m}L_{O_i}^y}{\sum_{i=0}^{m}L_{G_i}^y}
  \label{eq4}
\end{equation}

It is notable that we utilize the projection length on $x$ and $y$ axis rather than the bounding box area that is more commonly used in IoU ~\cite{rahman2016optimizing, giou}. This is to make the metric more robust against the equivalence of a single bounding box and its division of temporally consecutive bounding boxes mentioned above, which will be discussed later in Fig.4(d-e). 

Similarly, for the precision metric, we first calculate the overlapped regions between all the GT bounding boxes $\left \{G_i\right \}_{1,2,...,m}$ and each predicted bounding box $P_j$, 
\begin{equation}
O_j = \left \{ G_1 \cap  P_j, G_2 \cap  P_j, ... , G_n \cap  P_j \right \}
  \label{eq5}
\end{equation}
and then obtain the union frames $L_{O_j}^x$ and $L_{O_j}^y$ of all GT boxes for each predicted box. The precision is defined in a similar way as the recall:
\begin{equation}
Precision = \frac{\sum_{j=0}^{n}L_{O_j}^x}{\sum_{j=0}^{n}L_{P_j}^x}\cdot \frac{\sum_{j=0}^{n}L_{O_j}^y}{\sum_{j=0}^{n}L_{P_j}^y}
  \label{eq6}
\end{equation}
, where $L_{P_j}^x$ and $L_{P_j}^y$ are the width and height of predicted box $P_j$ respectively.

To calculate the final score, the harmonic mean of recall and precision, i.e. $Fscore$, is adopted as the primary metric:
\begin{equation}
Fscore = \frac{2\cdot Recall\cdot Precision}{Recall + Precision}
  \label{eq7}
\end{equation}

\subsection{Comparison}
Technically, previous segment and frame P/R protocol cannot be used for video pairs as input. To make them comparable to the proposed metric in this scenario, we extend them by simply calculating the metrics on both $x$ and $y$ axis respectively. Fig.4(c-f) provide several simplified and extreme cases with the evaluation results under different metrics. As shown in Fig.4(c), extended segment P/R metric cannot reflect the inaccurate boundary of the predicted copied segments. For multiple sub-segments emphasized in Sec.4.1, previous segment metric also lacks robustness indicated in Fig.4(d-e). Frame P/R shows poor measurements in the case shown in Fig.4(f) where predicted copied segments totally misalign with GT segments. 

By calculating the frame-level prediction accuracy inside each segment pair, our proposed metric shows better adaptability for different video copy scenarios. With IoU of the segment pair (represented as bounding box in similarity map) taken into account, the temporal correlation between two videos is highlighted in our metric. This overcomes the obvious drawback of frame-level statistic mentioned before but still maintains the accuracy measurement with fine granularity. Meanwhile, our metric is also robust for segment division that significantly affects previous segment metric. We also discuss the limitation of our proposed metric on some rare and extreme cases in Sec.S6 of Supplementary Material. In addition, mAP from action localization task ~\cite{caba2015activitynet, kay2017kinetics} is not suitable for copy localization. The core of mAP is to inspect the temporal (1D) IoU with the GT segment with evaluated result of true or false positive. However, the pair of input videos in video copy scenarios both might contain multiple partial copies, and the proposed metric should better measure prediction accuracy of copy overlapping segments between a video pair, rather than evaluating 1D temporal localization for a single video. 

\section{Benchmark}
\subsection{Pipeline}
We outline our pipeline in Fig.5. The pipeline starts with a pair of potential copied videos as input and then outputs the predicted copied segments. 

\begin{figure}[ht!] % use float package if you want it here
  \centering
  \includegraphics[width=1\linewidth]{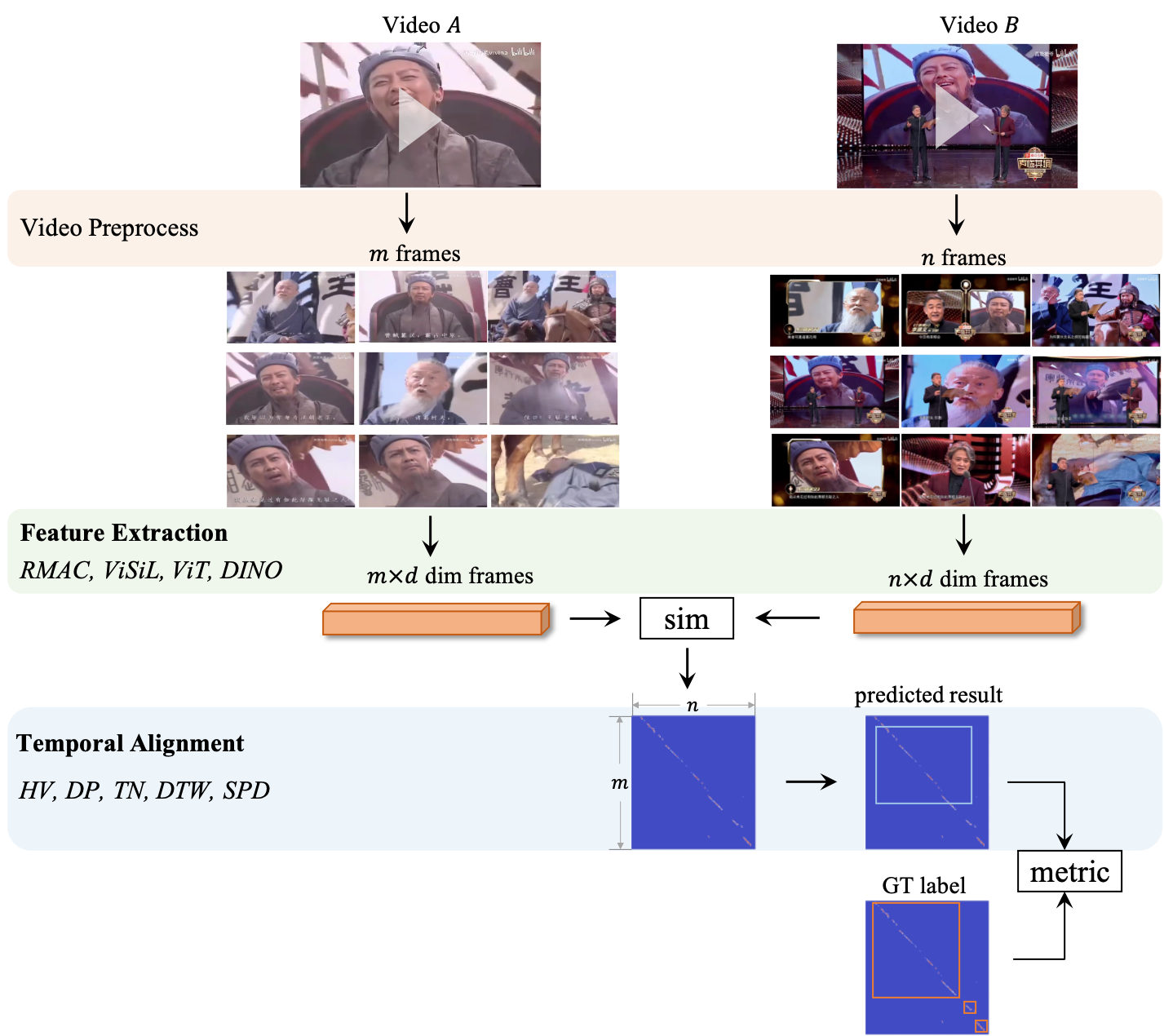}
  \caption{The benchmark pipeline.}
\end{figure}

The first step is the video preprocessing including video decoding and frame extraction. After this step, the input video is represented as a sequence of frames for subsequent process. Here, the frame extraction can be uniform or non-uniform based on keyframe selection or video summarization algorithms ~\cite{2021videosum,dsnet}. As this part is beyond the scope of this paper, we simply adopt the commonly used uniform sampling strategy.

After obtaining a set of frames (e.g., $m$ frames), we need to employ a visual descriptor on these $m$ frames and construct a $m\times d$ dimension embedding for each video. $d$ is the feature dimension for each frame. Then a frame-to-frame similarity map can be constructed by calculating the frame-to-frame similarity between the two video embeddings. In this paper, we select the following four feature extractors for comparison: classical and commonly used R-MAC feature ~\cite{mac17}; ViSiL feature that is proved SOTA on video-level retrieval task ~\cite{visil} but with high dimension; ViT that shows comparable performance on image retrieval task ~\cite{el2021training}; recently self-supervised pretrained model DINO ~\cite{dino} with competitive results on image copy detection. For R-MAC, ViT and DINO, similarity matrices are calculated by cosine similarity, and the ones for ViSiL feature are calculated by Chamfer Similarity which is consistent with ~\cite{visil}.

It is notable that all the selected feature extractors are frame-level. Meanwhile, there are methods ~\cite{I3D, C3D} can jointly learn spatial-temporal features. However, in our segment-level video copy detection task, the copy overlap boundaries should be precisely localized, hence the spatial-temporal features extracted on video clips with a minimum length of several seconds are not appropriate. But temporal info can be utilized after frame feature extraction and show improved performance on video retrieval ~\cite{tca}. We believe that both temporal and spatial attention can be flexibly inserted into the pipeline and the entire algorithm process could be trained in an end-to-end manner, and we leave this to our future work.

The final step is to find temporal alignments and output the copied segment pairs. As mentioned in Sec.4, the predicted results can be represented as bounding boxes in similarity map shown in Fig.5. Here, we re-implement the following five alignment methods: Hough Voting (HV) ~\cite{voting1}, Temporal Network (TN) ~\cite{temporaNetwork}, Dynamic Programming (DP) ~\cite{dp1}, dynamic time warping (DTW) ~\cite{dtw} and recent SPD \cite{spd}. The previous three methods are commonly used for video copy detection ~\cite{vcdb} and video retrieval task comparison ~\cite{visil}. DTW is usually adopted to match two time sequence, and we simply modify it for adaption to our task. SPD shows SOTA temporal alignment results on VCDB with the previous segment-level metric.

\subsection{Implementation Details}
\textbf{Dataset:} For the purpose of training and evaluation for benchmark and future works, we split VCSL dataset with at least one query set in train/val/test for each topic category. The query sets and copied video pairs in train/val/test are 60/32/30 and 97712/42031/27765 respectively. In addition, we supplement the same amount of negative samples (27765 pairs) to test set (55530 pairs in total) to evaluate the
algorithm performance more comprehensively. These negative pairs are constructed by random sampling from different video categories, which makes two videos irrelevant in each pair. The training, validation and test sets contain different sets of videos, providing completely independent contents. The detailed dataset split is given along with VCSL dataset release.

\textbf{Metric:} In Sec.4, we give the detailed evaluation metric for a copied video pair. For VCSL test datasets, we need to obtain an overall result on these 55530 pairs consisting of both positive and negative pairs. This overall metric can reflect segment-level alignment accuracy, and it is also influenced by video-level performance. In addition, we also utilize two auxiliary metrics, i.e., video-level False Rejection Rate (FRR)/ False Alarm Rate (FAR) performance and segment-level macro precision/recall performance on positive samples. In detail of the latter, we first obtain the average metric inside each query set of copied videos, and then calculate the macro average over query set as the overall result. This indicates the average segment detection accuracy of only positive samples (copied videos) in different categories.

\textbf{Feature extraction:} We extract one frame per second for each video. All models for frame feature are pretrained on ImageNet ~\cite{krizhevsky2012imagenet} without other external datasets.

\textbf{Temporal alignment:} Since all five temporal alignment algorithms are traditional methods without training process except for SPD, we tune their hyper parameters on the validation set of VCSL and compare the results on the test set. For SPD, we train two versions of the network respectively on VCDB and train/val set of VCSL, and also evaluate them on the test set of VCSL.

Other detailed experimental settings can be inferred in our released benchmark codes.

\subsection{Results and Discussion}
Table 2 presents the overall performance of all possible combinations of the features and alignment methods with our proposed metric on both positive and negative samples (55530 pairs). Table 3 shows video-level copy detection results regardless of copied localization accuracy. Table 4 gives segment-level macro evaluated results on only positive samples over query set (30 query sets with 27765 copied pairs).

\begin{table}[]
\caption{Comparison of benchmarked approaches with different features and alignment methods on both positive and negative samples. SPD is trained on train set of VCSL. (higher is better)}
\centering
\scalebox{1}{
\begin{tabular}{c|l|c|c|c}
\toprule
\multicolumn{2}{c|}{Overall benchmark}                                                & Recall & Precision & \textbf{F-score} \\ \toprule
\multirow{6}{*}{\begin{tabular}[c]{@{}c@{}}R-MAC\\ 512 dim.\end{tabular}}    & HV        & 57.46  & 41.28     & 48.04   \\ \cline{2-5} 
                                                                             & TN        & 64.38  & 54.03     & 58.75   \\ \cline{2-5} 
                                                                             & DP        & 40.03  & 61.92     & 48.62    \\ \cline{2-5} 
                                                                             & DTW       & 37.91  & 57.64     & 45.74    \\ \cline{2-5} 
                                                                             & SPD & 54.80  & 59.24     & 56.93   \\ \cline{2-5} 
                                        \midrule                                 
\multirow{6}{*}{\begin{tabular}[c]{@{}c@{}}ViSiL\\ 9*3840 dim.\end{tabular}} & HV        & 63.45  & 42.99     & 51.26   \\ \cline{2-5} 
                                                                             & TN        & 58.38  & 64.85     & 61.46   \\ \cline{2-5} 
                                                                             & DP        & 42.60  & 68.92     & 52.66   \\ \cline{2-5} 
                                                                             & DTW       & 29.34  & 69.63     & 41.28   \\ \cline{2-5} 
                                                                             & SPD & 56.56  & 60.59     & 58.50   \\ \cline{2-5} 
                                                                         \midrule
\multirow{6}{*}{\begin{tabular}[c]{@{}c@{}}ViT\\ 768 dim.\end{tabular}}      & HV        & 57.34  & 41.06     & 47.86   \\ \cline{2-5} 
                                                                             & TN        & 66.72  & 50.70     & 57.61   \\ \cline{2-5} 
                                                                             & DP        & 40.31  & 50.80     & 44.95   \\ \cline{2-5} 
                                                                             & DTW       & 40.00  & 42.08     & 41.01   \\ \cline{2-5} 
                                                                             & SPD & 51.50  & 60.36     & 55.58   \\ \cline{2-5} 
                                                                         \midrule
\multirow{6}{*}{\begin{tabular}[c]{@{}c@{}}DINO\\ 1536 dim.\end{tabular}}    & HV        & 66.99  & 39.83     & 49.96   \\ \cline{2-5} 
                                                                             & TN        & 77.61  & 48.89     & 59.99   \\ \cline{2-5} 
                                                                             & DP        & 45.75  & 64.87     & 53.66   \\ \cline{2-5} 
                                                                             & DTW       & 39.93  & 59.53     & 47.80   \\ \cline{2-5} 
                                                                             & SPD & 58.53  & 63.05     & 60.70   \\ \cline{2-5} 
                                                                         \bottomrule
\end{tabular}}
\end{table}

\begin{table}[]
\caption{Video-level copy detection results with different features and alignment methods on both positive and negative samples. SPD is trained on train set of VCSL. (lower is better)}
\centering
\scalebox{1}{
\begin{tabular}{c|l|c|c}
\toprule
\multicolumn{2}{c|}{Video-level performance}                                                & FRR & FAR  \\ \toprule
\multirow{5}{*}{\begin{tabular}[c]{@{}c@{}}R-MAC\\ 512 dim.\end{tabular}}    & HV        & 0.3614  & 0.0474        \\ \cline{2-4} 
                                                                             & TN        & 0.2286  & 0.0309        \\ \cline{2-4} 
                                                                             & DP        & 0.2464  & 0.0142        \\ \cline{2-4} 
                                                                             & DTW       & 0.4135  & 0.0028       \\ \cline{2-4} 
                                                                             & SPD & 0.3111  & 0.2036      \\ \cline{2-4} 
                                        \midrule                                 
\multirow{5}{*}{\begin{tabular}[c]{@{}c@{}}ViSiL\\ 9*3840 dim.\end{tabular}} & HV        & 0.3043  & 0.0012        \\ \cline{2-4} 
                                                                             & TN        & 0.2508  & 0.0059       \\ \cline{2-4} 
                                                                             & DP        & 0.2586  & 0.0002        \\ \cline{2-4} 
                                                                             & DTW       & 0.5689  & 0.0000       \\ \cline{2-4} 
                                                                             & SPD & 0.2868  & 0.1868        \\ \cline{2-4} 
                                                                         \midrule
\multirow{5}{*}{\begin{tabular}[c]{@{}c@{}}ViT\\ 768 dim.\end{tabular}}      & HV        & 0.3679  & 0.0329        \\ \cline{2-4} 
                                                                             & TN        & 0.1284  & 0.0976       \\ \cline{2-4} 
                                                                             & DP        & 0.1751  & 0.0943        \\ \cline{2-4} 
                                                                             & DTW       & 0.1879  & 0.0693        \\ \cline{2-4} 
                                                                             & SPD & 0.3547  & 0.1811       \\ \cline{2-4} 
                                                                         \midrule
\multirow{5}{*}{\begin{tabular}[c]{@{}c@{}}DINO\\ 1536 dim.\end{tabular}}    & HV        & 0.2773  & 0.0206        \\ \cline{2-4} 
                                                                             & TN        & 0.0685  & 0.1449        \\ \cline{2-4} 
                                                                             & DP        & 0.2085  & 0.0013        \\ \cline{2-4} 
                                                                             & DTW       & 0.4099  & 0.0003        \\ \cline{2-4} 
                                                                             & SPD & 0.2726  & 0.1732       \\ \cline{2-4} 
                                                                         \bottomrule
\end{tabular}}
\end{table}

% \usepackage{multirow}
\begin{table}[]
\caption{Segment-level macro evaluated results on only positive samples over query set. The footnote in SPD ($_1$) denotes SPD trained on VCDB. ($_2$) denotes SPD trained on train set of VCSL. (higher is better)}
\centering
\scalebox{0.85}{
\begin{tabular}{c|l|c|c|c}
\toprule
\multicolumn{2}{c|}{macro over Query Set}                                                & Recall & Precision & \textbf{F-score} \\ \toprule
\multirow{6}{*}{\begin{tabular}[c]{@{}c@{}}R-MAC\\ 512 dim.\end{tabular}}    & HV        & 77.65  & 75.08     & 76.34   \\ \cline{2-5} 
                                                                             & TN        & 82.05  & 87.95     & 84.90   \\ \cline{2-5} 
                                                                             & DP        & 61.04  & 87.37     & 71.87    \\ \cline{2-5} 
                                                                             & DTW       & 55.10  & 85.74     & 67.09    \\ \cline{2-5} 
                                                                             & SPD$_{1}$ & 79.39  & 91.37     & 84.96   \\ \cline{2-5} 
                                                                             & SPD$_{2}$ & 82.16  & 89.79     & {85.81}   \\ \cline{2-5} \midrule
\multirow{6}{*}{\begin{tabular}[c]{@{}c@{}}ViSiL\\ 9*3840 dim.\end{tabular}} & HV        & 81.93  & 71.64     & 76.44   \\ \cline{2-5} 
                                                                             & TN        & 82.16  & 89.56     & 85.70   \\ \cline{2-5} 
                                                                             & DP        & 64.28  & 89.76     & 74.91   \\ \cline{2-5} 
                                                                             & DTW       & 54.27  & 91.40     & 68.10   \\ \cline{2-5} 
                                                                             & SPD$_{1}$ & 79.16  & 91.45     & 84.86   \\ \cline{2-5} 
                                                                             & SPD$_{2}$ & 83.87  & 88.97     & {86.34}   \\ \cline{2-5} \midrule
\multirow{6}{*}{\begin{tabular}[c]{@{}c@{}}ViT\\ 768 dim.\end{tabular}}      & HV        & 76.71  & 75.70     & 76.20   \\ \cline{2-5} 
                                                                             & TN        & 83.60  & 86.22     & 84.89   \\ \cline{2-5} 
                                                                             & DP        & 60.61  & 81.20     & 69.41   \\ \cline{2-5} 
                                                                             & DTW       & 55.40  & 72.99     & 62.99   \\ \cline{2-5} 
                                                                             & SPD$_{1}$ & 80.56  & 90.28     & 85.14   \\ \cline{2-5} 
                                                                             & SPD$_{2}$ & 81.61  & 90.94     & {86.02}   \\ \cline{2-5} \midrule
\multirow{6}{*}{\begin{tabular}[c]{@{}c@{}}DINO\\ 1536 dim.\end{tabular}}    & HV        & 81.46  & 73.17     & 77.09   \\ \cline{2-5} 
                                                                             & TN        & 88.74  & 83.69     & 86.14   \\ \cline{2-5} 
                                                                             & DP        & 64.36  & 86.58     & 73.83   \\ \cline{2-5} 
                                                                             & DTW       & 57.16  & 84.92     & 68.33   \\ \cline{2-5} 
                                                                             & SPD$_{1}$ & 81.23  & 90.66     & 85.69   \\ \cline{2-5} 
                                                                             & SPD$_{2}$ & 84.67  & 90.31     & {87.40}   \\ \cline{2-5} \bottomrule
\end{tabular}}
\end{table}

As can be shown from Table 2-4, the frame feature does indeed have impacts on the performance, and there shows similar trend with different alignment methods. DINO achieves better results with a moderate size of feature dimension, which might be attributed to the transformer architecture and the self-supervised framework. However, the effect of features on the final results is not as dramatic as expected, especially considering different feature dimensions. By observing the similarity map, it can be found that the pattern of copied segments is not obvious for some hard cases (picture in picture from variety shows, crop in a large margin from kichiku), and all methods show poor performance. Similarity maps on some example hard cases are given in Sec.S3 of Supplementary Material. We suspect that it is due to the limitation of global features that can not capture local correspondences between severely transformed frames. We hope that the hard cases in VCSL give some insights to develop more powerful feature representations for the segment-level video copy detection task.

In the aspect of temporal alignment methods, SPD and TN perform better than other methods on overall benchmark in Table 2. If we focus on copy location accuracy on only positive samples, DINO + SPD achieve highest F-score of over 87\% shown in Table 4. However, if we dive into the detailed results on video-level in Table 3, FAR/FRR of SPD are unsatisfactory with higher results than other alignment methods. Therefore, SPD can accurately localize copied segments on positive samples, but it is not suitable for only video-level copy detection. In Table.4, it is notable that SPD results trained on VCDB dataset are lower than SPD with VCSL, and this indicates the importance of large-scale and well-annotated datasets, especially for supervised learning methods. Moreover, results on some specific query sets are only around 50\% which are far from satisfactory, especially on some query sets in kichiku and movie category with significant temporal and spatial editing. Such recently emerged copy infringement types in VCSL bring great challenges to these temporal alignment methods designed for near duplicated cases. The details and bad cases analysis are given in Sec.S4 of Supplementary Material. 

Besides the overall results above, we also evaluate the algorithms at different data distribution with finer granularity. Corresponding to Fig.2, F-score performance results on video duration, segment duration, segment number per video pair, copy duration percentage are indicated in Sec.S5 of Supplementary Material with detailed analysis. Video pairs containing more segment copies and lower copy duration percentage meet significantly more difficulties with lower results. VCSL provides a large amount of these types to motivate future algorithm evolution. In addition, we can also observe that temporal alignment methods show different adaptability on various data distribution and situations, e.g., SPD shows better performance on backwards-running videos, and TN is more suitable for multiple copied segments per video pair. This is due to their different definition and constraints on copy detection task. Fine-grained video copy detection and consequent model fusion might also be opportunities for future research.

\section{Conclusion}
This work represents the currently largest segment-level video copy detection dataset, VCSL. Compared with the existing partial copy detection dataset (VCDB), VCSL has two orders of magnitude more labelled data, and is collected from realistic, challenging YouTube and Bilibili videos. In addition, we refine the evaluation protocol and jointly propose a new metric to address existing problems revealed by the previous evaluation protocol and metric. Four feature extraction methods and five temporal alignment methods are quantitatively evaluated and compared, which reveals interesting future research directions. We hope that the public availability of VCSL and new thoughtful metric will motivate even more interest in this important and applicable field for video copy detection and copyright protection.
{\small
\bibliography{egbib}
}

\end{document}

\title{A Large-scale Comprehensive Dataset and Copy-overlap Aware Evaluation Protocol for Segment-level Video Copy Detection: Supplementary Material}

\maketitle

%%%%%%%%% BODY TEXT
\renewcommand{\thesection}{S\arabic{section}}
\section{Frame-to-frame similarity map}

The frame-to-frame similarity matrix based on frame-level features is capable of representing temporal structure between the copied videos and it has proved to be successful to learn similarity patterns of the pairwise frame similarities. 

The matrix element $s_{ij}$ is the similarity score between a frame feature $f_i$ of video $A$ and $f'_j$ of a potential copied video $B$. 
\begin{equation}
\label{eq1}
  s_{ij}= \frac{f_i * f'_j}{||f_i||*||f'_j||}
\end{equation}
where $i$,$j$ indicates the $i$th frame of video $A$ and the $j$th frame of the video $B$. An example of frame-to-frame similarity is shown in Figure 1. The red bounding box indicated to temporal copied part of two original videos is a copied segment pair. This pattern in the bounding box which is similar to an oblique straight line represents the temporal sequential copy between two videos with high similarity scores between the temporal corresponding frames.

\begin{figure}[ht!] % use float package if you want it here
  \centering
  \includegraphics[width=0.5\linewidth]{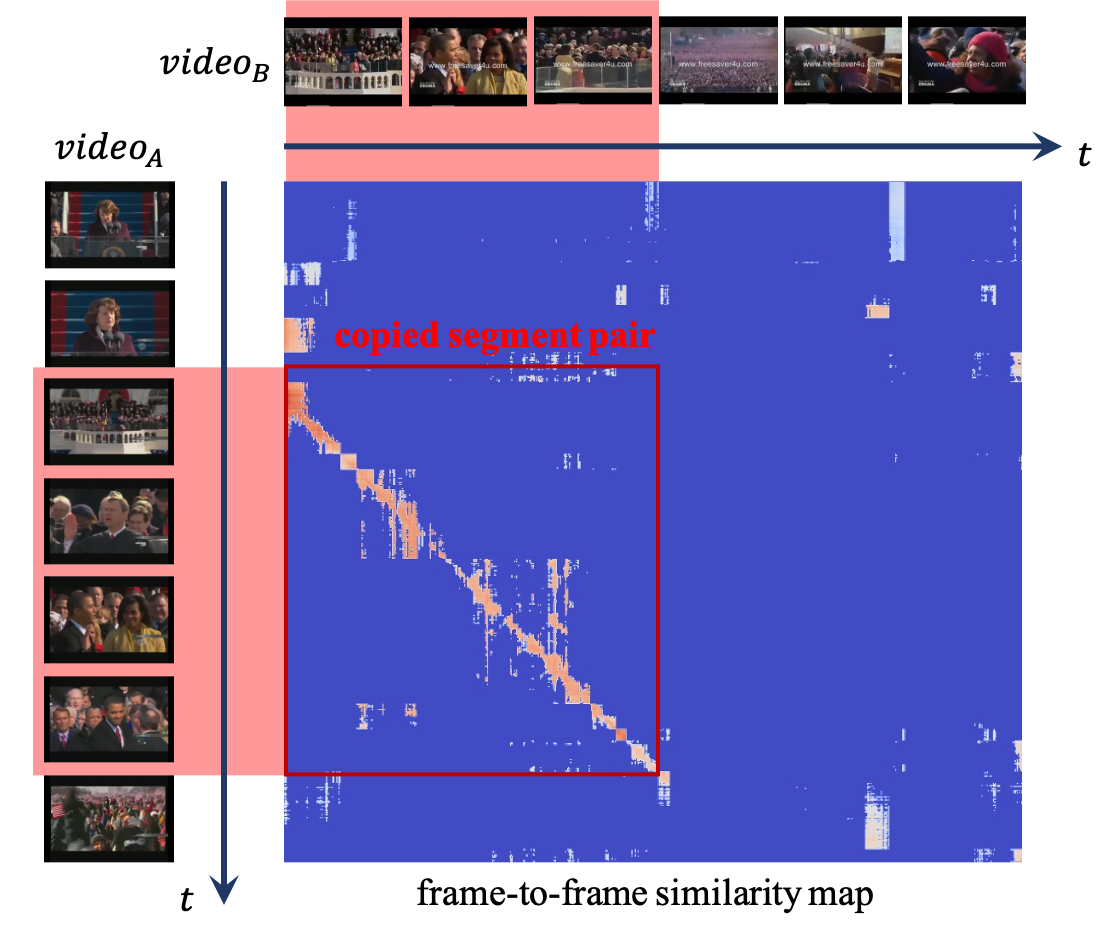}
  \caption{An frame-to-frame similarity map example.}
\end{figure}

\section{Example copied videos in VCSL}
As we mentioned in Section 3.2 in our main text, VCSL covers lots of realistic spatial and temporal transformations. Due to the representation restriction by only figures and text descriptions in this manuscript, we only list some frame transformations (spatial transformation) here. We recommend readers to discover the various temporal transformations in VCSL covering reverse, loop, video mushup, acceleration and deceleration. Moreover, we have also tried to label different types of transformations on VCSL. But both the spatial and temporal transformations are too various to be concluded by limited types, and we finally give up this plan.

Fig.2 below shows some typical spatial transformations in VCSL including crop, filter, text overlay, background, cam-cording, picture in picture, even recent deepfake, etc. The videos marked with the same background blue box are from the same query set, and the first video inside each background box is the seed video. There are a wide range of content transformations among over 280k segment copies in VCSL, and these realistic skillful transformations bring great challenges to segment-level copy detection.

\begin{figure}[ht!] % use float package if you want it here
  \centering
  \includegraphics[width=\linewidth]{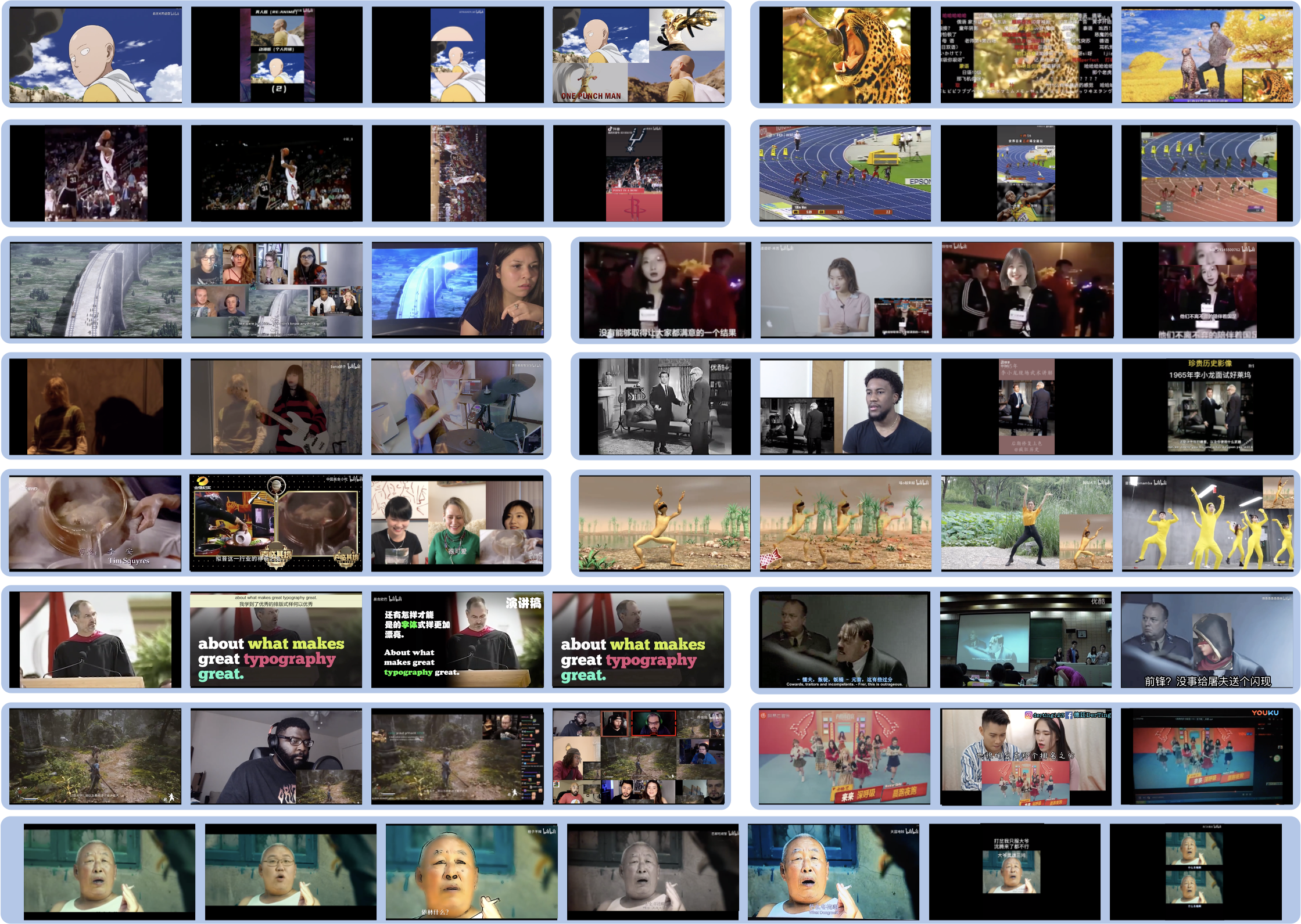}
  \caption{Example copied videos from VCSL. The videos with the same background blue box are from the same query set, and the first video of each background box is seed video.}
\end{figure}

\section{Similarity maps on hard cases}
In Section 5.3 of our main text, we observe that the effect of features on the final results is not as dramatic as expected, especially considering different feature dimensions. By observing the similarity map, it can be found that the patterns of these copied segments are not obvious and show no contrast with the surroundings for some hard cases. Similarity maps on some example hard cases including picture in picture, severe crop and cam-cording are given in Fig.3. 

\begin{figure}[ht!] % use float package if you want it here
  \centering
  \includegraphics[width=0.85\linewidth]{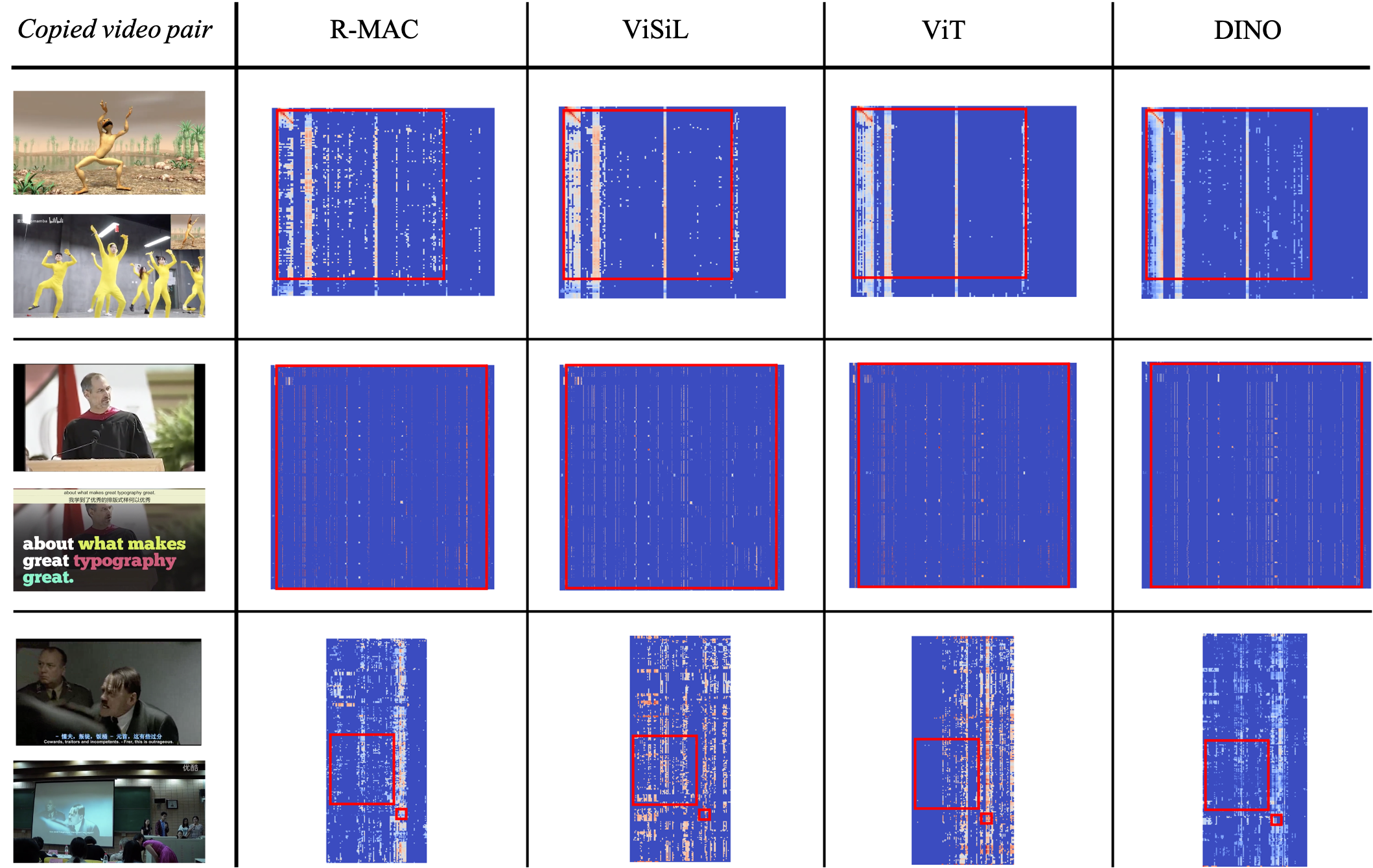}
  \caption{Similarity maps of different features on some hard cases. Red bounding boxes indicate ground-truth copied segment locations. The left column shows the original copied video pairs. Almost all the experimental feature extractors cannot work well on these hard cases.}
\end{figure}

The red bounding boxes in Fig.3 indicate the manually annotated copied locations. Theoretically, there should be distinct copied patterns at the ground truth copied locations (e.g., pattern shown in Fig.1) if the frame feature extractor works ideally. However, the unrecognizable copied patterns in Fig.3 show the limitation of current global features that they cannot extract the similarity between the temporal corresponding frames. We hope that the hard cases in VCSL give some insights to develop more powerful feature representations for the segment-level video copy detection task.

\section{Benchmark results at different topic category}

We show the overall results of all combinations of feature extractors and temporal alignment methods in Section 5.3 of our main text. In this part of Supplementary Material, more fine-grained results are calculated by averaging the metric results of each video pair in different topic categories. The detailed F-score results are shown in Table.1 above with bold text highlighted on the best performance among all the temporal aligned methods. The frame feature extractor here is DINO with better feature representation verified before.

As we mentioned in our main text, results on some specific query sets are only around 50\% which are far from satisfactory, especially on some query sets in kichiku and movie category with significant temporal and spatial editing. Among all the temporal alignment algorithms, SPD trained on VCSL achieves the best accuracy results on more than half of the topic categories (6/11), and TN also has three highest scores but only slightly better than SPD$_2$ on these three categories. It is interesting that the simplest HV method obtains the best result on the most difficult topic kichiku, even though this best result is also less than 60\%. Besides the impact of feature extractors, all these temporal alignment methods also need to be specifically optimized for this recently emerged copy infringement types in VCSL.

\begin{table}[]
\caption{Benchmark F-score results at different topic category}
\centering
\begin{tabular}{@{}c|m{1cm}<{\centering}|m{1cm}<{\centering}|m{1cm}<{\centering}|m{1cm}<{\centering}|m{1cm}<{\centering}|m{1cm}<{\centering}|m{1cm}<{\centering}|m{1cm}<{\centering}|m{1cm}<{\centering}|m{1cm}<{\centering}|m{1cm}<{\centering}@{}}
\toprule
Method & variety show      & games      & music videos     & news      &  sports      & daily life    & kichiku      & adver-tisement      & anima-tion      & movies      & TV series     \\ \midrule
HV     & 81.33 & 93.34 & 67.20 & 77.48 & 53.57 & 47.33 & \textbf{59.96} & 67.42 & 90.67 & 58.90 & 91.16 \\ \hline
TN   & \textbf{97.14} & \textbf{97.41} & 76.88 & 89.69 & 59.38 & 65.52 & 57.19 & 75.36 & 91.43 & 65.41 & \textbf{95.68} \\\hline
DP   & 81.97 & 92.02 & 71.87 & 83.10 & 57.10 & 37.23 & 46.79 & 71.13 & 86.35 & 60.37 & 90.26 \\\hline
DTW  & 76.87 & 92.75 & 65.21 & 72.48 & 48.98 & 45.06 & 53.75 & 61.30 & 92.82 & 48.07 & 84.67 \\\hline
SPD$_1$   & 96.09 & 94.69 & 86.52 & 89.34 & 58.23 & 68.37 & 49.77 & \textbf{76.39} & 93.92 & 55.10 & 93.96 \\\hline
SPD$_2$   & 96.04 & 96.49 & \textbf{89.27} & \textbf{92.84} & \textbf{70.79} & \textbf{71.10} & 54.37 & 74.93 & \textbf{96.92} & \textbf{70.18} & 95.46 \\ \bottomrule
\end{tabular}
\end{table}

\section{Benchmark results at different data distributions}

In this section, we evaluate the algorithms at different data distributions corresponding to statistics in Fig.2 of our main text. In detail, F-score performance results on video duration, segment duration, segment number per video pair, copy duration percentage are indicated in Table 2-5 above. Here, the video and segment duration are calculated by averaging the two videos and segments duration of each pair. Similar with Table 1, frame feature extractor here is also DINO.

\begin{table}[]
\caption{Benchmark F-score results at different video durations}
\centering
\begin{tabular}{@{}c|m{1.5cm}<{\centering}|m{1.5cm}<{\centering}|m{1.5cm}<{\centering}|m{1.5cm}<{\centering}|m{1.5cm}<{\centering}|m{1.5cm}<{\centering}|m{1.5cm}<{\centering}|m{1.5cm}<{\centering}}\toprule
Method & 10-30s  & 30-60s  & 60s-2min & 2-4min  & 4-8min  & 8-15min & 15-30min & \textgreater{}30min \\\midrule
HV     & 66.15 & 64.09 & 59.30  & 62.87 & 59.59 & 51.75 & 75.50  & 59.66             \\\hline
TN   & \textbf{83.45} & \textbf{68.67} & 60.90  & 64.79 & 70.98 & 62.45 & 90.45  & 69.89             \\\hline
DP   & 78.04 & 62.70 & 53.11  & 54.51 & 61.10 & 52.51 & 71.58  & 56.81             \\\hline
DTW  & 76.11 & 55.98 & 51.75  & 50.33 & 54.09 & 41.77 & 55.61  & 54.57             \\\hline
SPD$_1$   & 75.68 & 63.01 & 55.08  & 60.56 & 63.33 & 65.15 & 88.46  & 69.67             \\\hline
SPD$_2$   & 77.40 & 67.24 & \textbf{63.54}  & \textbf{71.78} & \textbf{76.47} & \textbf{68.07} & \textbf{90.68}  & \textbf{71.82}    \\ \bottomrule       
\end{tabular}
\end{table}

\begin{table}[]
\caption{Benchmark F-score results at different segment durations}
\centering
\begin{tabular}{@{}c|m{1.35cm}<{\centering}|m{1.35cm}<{\centering}|m{1.35cm}<{\centering}|m{1.35cm}<{\centering}|m{1.35cm}<{\centering}|m{1.35cm}<{\centering}|m{1.35cm}<{\centering}|m{1.35cm}<{\centering}|m{1.35cm}<{\centering}}
\toprule
Method & 0-5s    & 5-10s   & 10-20s  & 20-45s  & 45-90s  & 90s-3min & 3-6min  & 6-30min & \textgreater{}30min \\ \midrule
HV     & \textbf{45.76} & \textbf{48.18} & 50.12 & 59.26 & 65.60 & 74.59  & 86.44 & 79.64 & \textbf{90.19}             \\\hline
TN   & 17.50 & 31.01 & 49.91 & 64.81 & \textbf{71.80} & 79.10  & 88.58 & 67.44 & 77.06             \\\hline
DP   & 19.49 & 35.39 & 51.50 & 56.07 & 52.95 & 64.73  & 83.86 & 68.98 & 76.45             \\\hline
DTW  & 28.61 & 30.24 & 40.84 & 51.96 & 56.44 & 69.63  & 82.84 & 67.06 & 76.45             \\\hline
SPD$_1$   & 9.99  & 18.57 & 38.32 & 61.30 & 69.38 & 83.86  & \textbf{92.73} & \textbf{79.71} & 76.09             \\\hline
SPD$_2$   & 25.87 & 43.65 & \textbf{63.85} & \textbf{65.12} & 68.47 & \textbf{85.53}  & 92.52 & 77.31 & 75.84             \\ \bottomrule
\end{tabular}
\end{table}

\begin{table}[]
\caption{Benchmark F-score results at different segment numbers per video pair}
\centering
\begin{tabular}{@{}c|m{2cm}<{\centering}|m{2cm}<{\centering}|m{2cm}<{\centering}|m{2cm}<{\centering}|m{2cm}<{\centering}|m{2cm}<{\centering}}
\toprule
Method & 1       & 2       & 3       & 4       & 5       & \textgreater{}5 \\ \midrule
HV     & 65.31 & 52.55 & \textbf{45.83} & \textbf{52.13} & \textbf{42.24} & 49.25         \\\hline
TN   & 70.63 & 51.46 & 44.75 & 45.76 & 37.28 & \textbf{51.68}         \\\hline
DP   & 60.91 & 45.66 & 36.19 & 35.26 & 25.72 & 36.75         \\\hline
DTW  & 55.79 & 41.99 & 44.79 & 43.96 & 41.95 & 36.49         \\\hline
SPD$_1$   & 66.45 & 41.28 & 27.76 & 35.91 & 19.95 & 35.90         \\\hline
SPD$_2$   & \textbf{74.31} & \textbf{56.55} & 40.16 & 42.38 & 18.93 & 36.58         \\ \bottomrule
\end{tabular}
\end{table}

\begin{table}[]
\caption{Benchmark F-score results at different copy duration percentages}
\centering
\begin{tabular}{@{}c|m{2cm}<{\centering}|m{2cm}<{\centering}|m{2cm}<{\centering}|m{2cm}<{\centering}|m{2cm}<{\centering}}
\toprule
Method & 0-20\%  & 20-40\% & 40-60\% & 60-80\% & 80-100\% \\ \midrule
HV     & 43.67 & 53.70 & 66.02 & 67.99 & 79.28  \\\hline
TN   & 43.19 & 55.86 & 70.39 & \textbf{75.06} & 84.14  \\\hline
DP   & 45.09 & 53.70 & 57.65 & 57.16 & 73.58  \\\hline
DTW  & 35.19 & 47.77 & 49.92 & 59.83 & 79.50  \\\hline
SPD$_1$   & 22.05 & 54.87 & \textbf{71.26} & 69.24 & 87.68  \\\hline
SPD$_2$   & \textbf{59.63} & \textbf{62.95} & 69.19 & 70.77 & \textbf{88.30}  \\ \bottomrule
\end{tabular}
\end{table}

As can be shown in Table 2, SPD$_2$ (trained on VCSL) outperforms others on video duration longer than 1 min, and TN performs better on short videos. This is due to the resize operation in the preprocess of SPD, and this significantly rescales the similarity maps of short video pairs. In contrast, TN and other traditional methods without training process directly operate on the frame similarities, and this is more suitable for short videos. Table 3 shows the benchmark results of different segment durations, and SPD achieves better performance on most of the duration ranges with larger amounts in VCSL (amount data on segment duration can be referred from Fig.2(b) in the main text). For the extreme short segments (\textless{}10s) and long segments (\textgreater{}30min), HV method obtains the best result. After manually going through these segments, it is found that entire copied segments covering the diagonal of similarity map occupy most in these extreme cases and HV is suitable for these cases. From Table 4 and Table 5, video pairs containing more segment copies and lower copy duration percentages meet significantly more difficulties with lower results. In Table 5, SPD significantly outperforms other methods on copy duration percentage with larger quantity of video pair data (0-20\% and 80-100\%, indicated from Fig.2(d) in the main text). This might also be attributed to the scale and diversity of training dataset of VCSL. Overall, large-scale and well-annotated datasets are essential for supervised learning methods, and temporal alignment methods show different adaptability on various data distributions and situations.

\section{Limitation of metric on extreme cases}

We propose our new metric in Section.4 of the main text, and this metric fully considers the segment division equivalence of copy detection task. But this consideration also brings limitations on some rare and extreme cases shown in Fig.4 below. As we mentioned after Eq.(4) of main text, we utilize the projection length on $x$ and $y$ axis rather than the bounding box area that is more commonly used in IoU. This is to make the metric more robust against the equivalence of a single bounding box and its division of temporally consecutive bounding boxes shown in Fig.4(a). However, the wrong prediction with reversed order can also be measured with high scores shown in Fig.4(b), and extra prediction boxes inside ground truth box are also not punished by our metric shown in Fig.4(c). This is the trade-off between robustness on segment division and discrimination on inner inaccurate prediction. This trade-off cannot be perfectly solved since even the most precise annotation can only be finished at segment-level (boxes in similarity map) rather than frame-level (points in the similarity map) while considering annotation cost and operability. After observation on the prediction results, we find that the cases similar with Fig.4(a) appear far more frequently than Fig.4(b) and Fig.4(c). Similarity maps with a large GT box containing several inaccurate predicted boxes rarely happen with an optimized temporal alignment method. Therefore, we have to sacrifice the measurement accuracy on the rare cases like Fig.4(b) and (c), and make our metric to be robust against common segment division shown in Fig.4(a).

\begin{figure}[ht!] % use float package if you want it here
  \centering
  \includegraphics[width=0.9\linewidth]{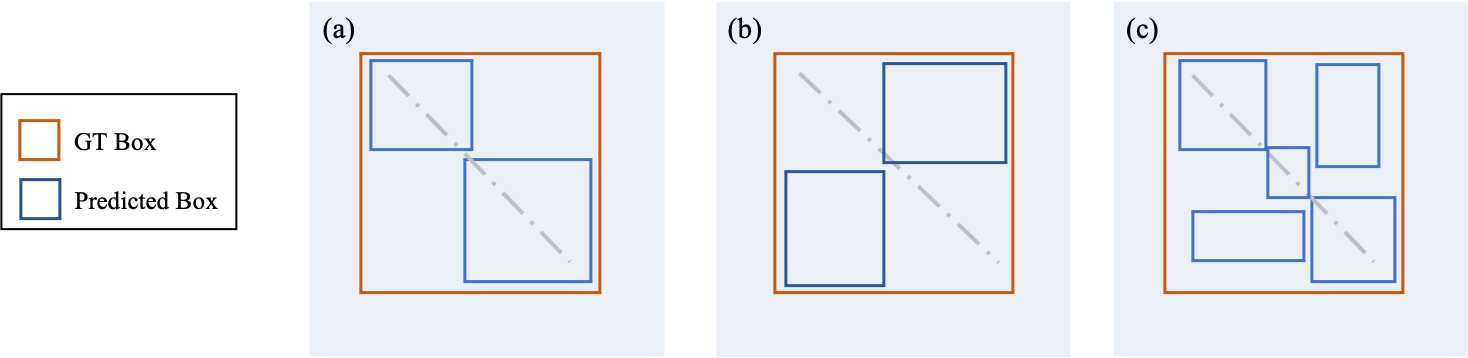}
  \caption{All the above situations are evaluated with high precision and recall results by our metric. However, (b) and (c) predict inaccurate localization. In detail, (a)correct measurement of our metric considering segment division equivalence (same with Fig.4(d) in main text); (b)inaccurate measurement with reversed-order predicted boxes; (c)inaccurate measurement with extra wrong predicted boxes.}
\end{figure}

{\small
\bibliographystyle{ieee_fullname}
\bibliography{egbib}
}